\documentclass[final]{cvpr}

\usepackage{times}
\usepackage{epsfig}
\usepackage{graphicx}
\usepackage{subfigure}
\usepackage{caption}
\usepackage{amsmath}
\usepackage{amssymb}
\usepackage{multirow}
\usepackage{color}
\usepackage{booktabs}
\usepackage{bbding}
\usepackage{makecell}

\usepackage[pagebackref,breaklinks,colorlinks]{hyperref}

\usepackage[capitalize]{cleveref}
\crefname{section}{Sec.}{Secs.}
\Crefname{section}{Section}{Sections}
\Crefname{table}{Table}{Tables}
\crefname{table}{Tab.}{Tabs.}

\graphicspath{{./figs/}{./figs/result/}}

\begin{document}



\title{PP-YOLOE: An evolved version of YOLO}


\author{Shangliang Xu, Xinxin Wang, Wenyu Lv, Qinyao Chang, Cheng Cui, \\
Kaipeng Deng, Guanzhong Wang, Qingqing Dang, Shengyu Wei, Yuning Du, Baohua Lai \\
Baidu Inc.\\
\tt\small \{xushangliang, wangxinxin08, lvwenyu01, dengkaipeng, dangqingqing\} @baidu.com
}

\maketitle

\begin{abstract}

  In this report, we present PP-YOLOE, an industrial state-of-the-art object detector with high performance and friendly deployment. We optimize on the basis of the previous PP-YOLOv2, using anchor-free paradigm, more powerful backbone and neck equipped with CSPRepResStage, ET-head and dynamic label assignment algorithm TAL. We provide s/m/l/x models for different practice scenarios. As a result, PP-YOLOE-l achieves \textbf{51.4 mAP} on COCO test-dev and \textbf{78.1 FPS} on Tesla V100, yielding a  remarkable improvement of (\textbf{+1.9 AP}, \textbf{+13.35\% speed up}) and (\textbf{+1.3 AP}, \textbf{+24.96\% speed up}), compared to the previous state-of-the-art industrial models PP-YOLOv2 and YOLOX respectively. Further, PP-YOLOE inference speed achieves \textbf{149.2 FPS} with TensorRT and FP16-precision. We also conduct extensive experiments to verify the effectiveness of our designs. Source code and pre-trained models are available at PaddleDetection$\footnote{\scriptsize\url{https://github.com/PaddlePaddle/PaddleDetection}\label{ppdet}}$. 
  
\end{abstract}


\section{Introduction}

One-stage object detector is popular in real-time applications due to excellent speed and accuracy trade-off. The most prominent architecture among one-stage detectors is the YOLO series\cite{redmon2016you, redmon2017yolo9000, redmon2018yolov3, bochkovskiy2020yolov4, wang2021scaledyolov4, glenn_jocher_2022_6222936, ge2021yolox, long2020pp, huang2021pp-yolov2}. Since YOLOv1\cite{redmon2016you}, YOLO series object detectors have undergone tremendous changes in network structure, label assignment and so on. At present, YOLOX\cite{ge2021yolox} achieves an optimal balance of speed and accuracy with 50.1 mAP at the speed of 68.9 FPS on Tesla V100.

\begin{figure}[ht]
\centering
\includegraphics[width=\linewidth]{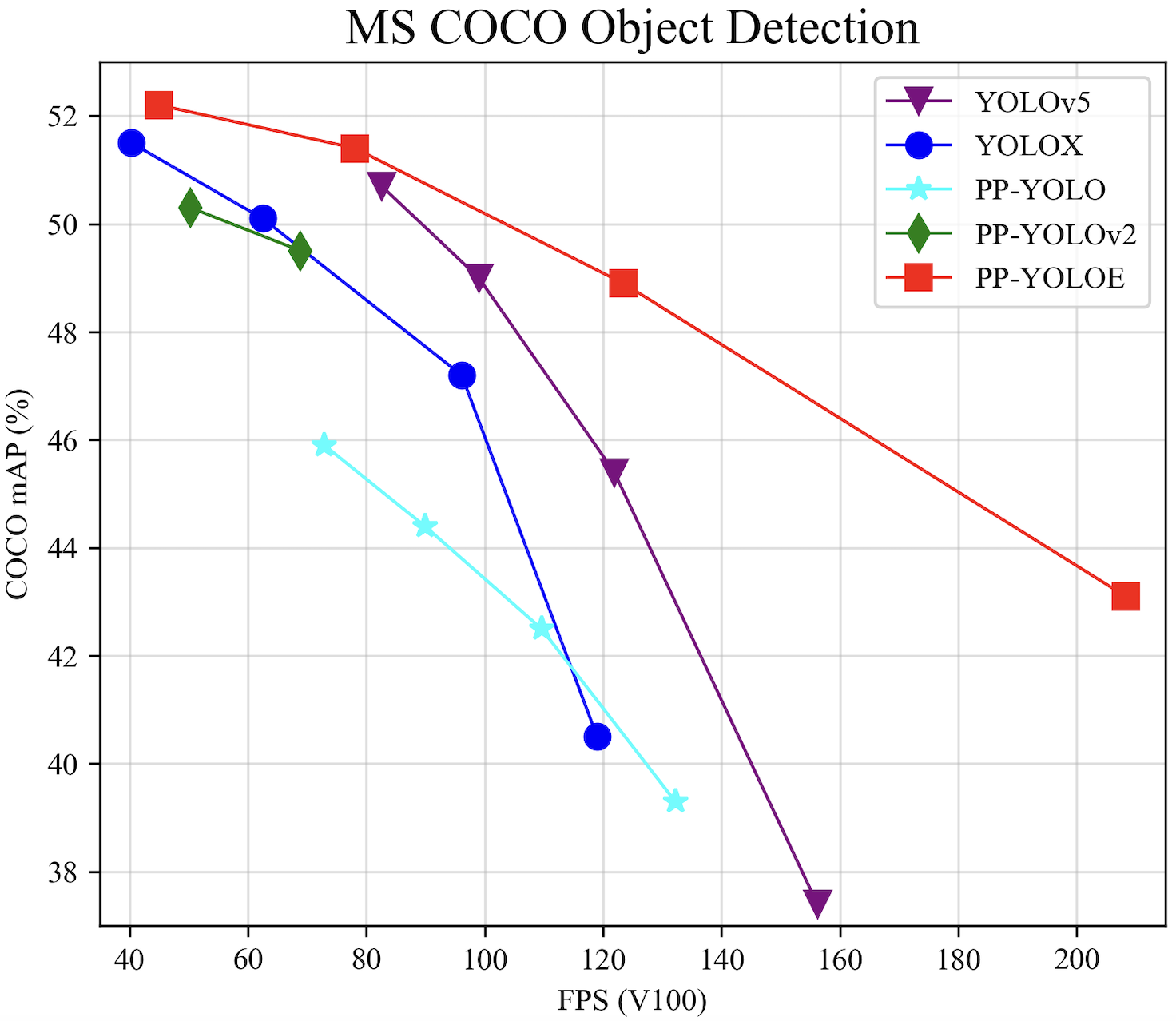} 
\caption{Comparison of the PP-YOLOE and other state-of-the-art models. PP-YOLOE-l achieves 51.4 mAP on COCO \textit{test-dev} and  78.1 FPS on Tesla V100, obtains 1.9 AP and 9.2 FPS improvement compared with PP-YOLOv2\cite{huang2021pp-yolov2}. }
\label{figure1}
\end{figure}

YOLOX introduces advanced anchor-free method equipped with dynamic label assignment to improve the performance of detector, significantly outperforming YOLOv5\cite{glenn_jocher_2022_6222936} in terms of precision. Inspired  by YOLOX, we further optimize our previous work PP-YOLOv2\cite{huang2021pp-yolov2}. PP-YOLOv2 is a high-performance one-stage detector with 49.5 mAP at the speed of 68.9 FPS on Tesla V100. Based on PP-YOLOv2, we proposed an evolved version of YOLO named PP-YOLOE. PP-YOLOE avoids using operators like deformable convolution\cite{dai2017deformable, zhu2018deformable} and Matrix NMS\cite{wang2020solov2} to be well supported on various hardware. Moreover, PP-YOLOE can easily scale to a series of models for various hardware with different computing power. These characteristics further promote the application of PP-YOLOE in a wider range of practical scenarios.

As shown in \cref{figure1}, PP-YOLOE outperforms YOLOv5 and YOLOX in terms of speed and accuracy trade-off. Specifically, PP-YOLOE-l achieves 51.4 mAP on COCO with 640 $\times$ 640 resolution at the speed of 78.1 FPS, surpassing PP-YOLOv2 by 1.9\% AP and YOLOX-l by 1.3\% AP. Moreover, PP-YOLOE has a series of models, which can be simply configured through width multiplier and depth multiplier like YOLOv5. Our code has released on PaddleDetection\cite{ppdet2021}, with TensorRT and ONNX supported.


\begin{figure*}[ht]
	\centering
	\includegraphics[width=\textwidth]{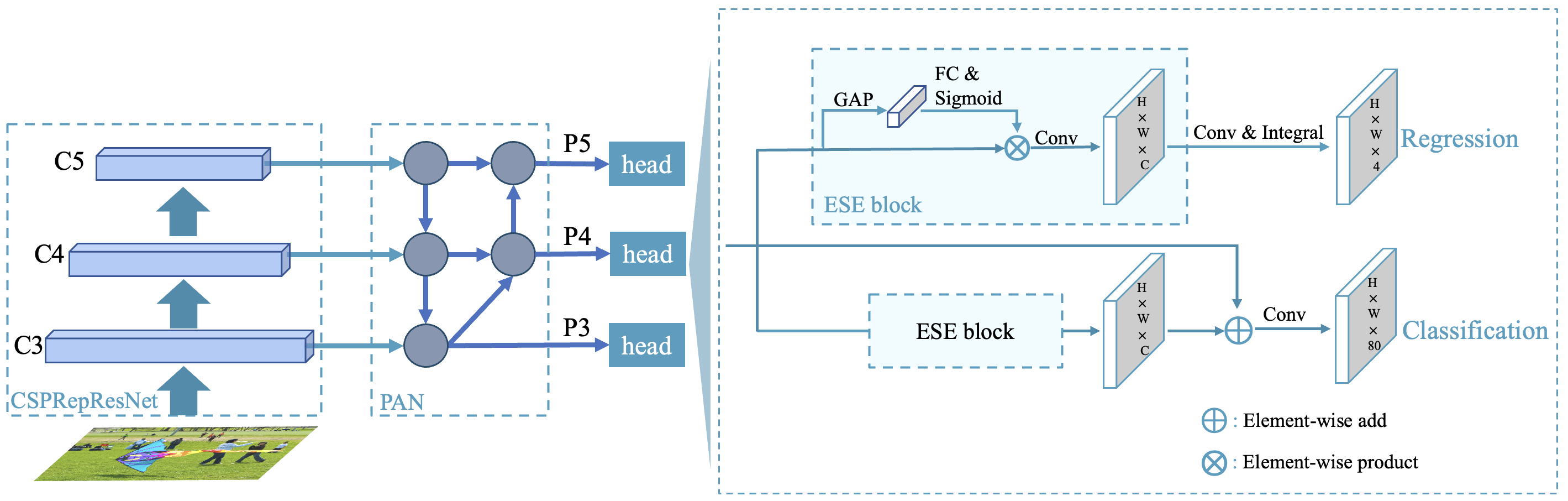}
	\caption{The model architecture of our PP-YOLOE. The backbone is CSPRepResNet, the neck is Path Aggregation Network (PAN), and the head is Efficient Task-aligned Head (ET-head). }
	\label{head_fig}
\end{figure*}

\section{Method}
In this section, we will first review our baseline model and then introduce the design of PP-YOLOE (\cref{head_fig}) in detail from the aspects of network structure, label assignment strategy, head structure and loss function.


\begin{figure*}[ht]
	\centering
	\subfigure[Simplified TreeBlock]{
	    \begin{minipage}[t]{0.22\linewidth}
			\centering
			\includegraphics[scale=0.55]{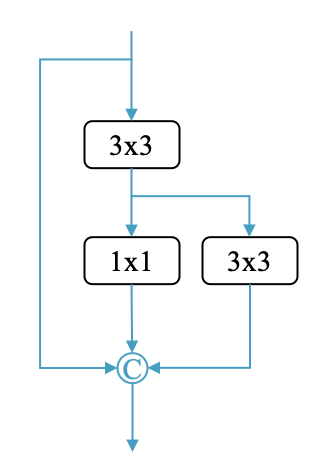}
			\label{fig:figure2a}
		\end{minipage}
	}
	\subfigure[Our RepResBlock during training]{
	    \begin{minipage}[t]{0.25\linewidth}
			\centering
			\includegraphics[scale=0.55]{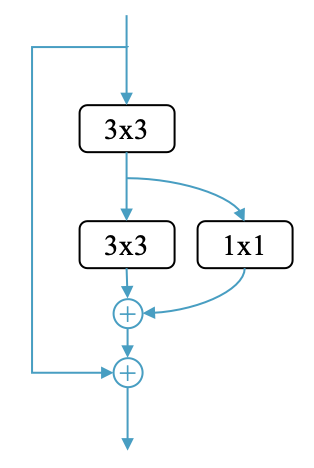}
			\label{fig:figure2b}
		\end{minipage}
	}
	\subfigure[Our RepResBlock during inference]{
	    \begin{minipage}[t]{0.25\linewidth}
			\centering
			\includegraphics[scale=0.55]{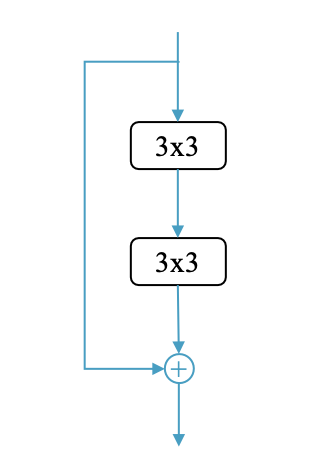}
			\label{fig:figure2c}
		\end{minipage}
	}
	\subfigure[Our CSPRepResStage]{
	    \begin{minipage}[t]{0.22\linewidth}
			\centering
			\includegraphics[scale=0.48]{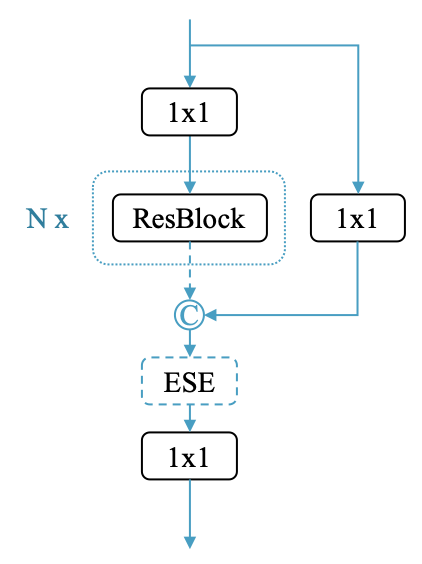}
			\label{fig:figure2d}
		\end{minipage}
	}
	\label{fig:figure2}
	\caption{Structure of our RepResBlock and CSPRepResStage}
\end{figure*}

\subsection{A Brief Review of PP-YOLOv2}

The overall architecture of PP-YOLOv2 contains the backbone of ResNet50-vd\cite{he2019bag} with deformable convolution\cite{zhu2018deformable}, the neck of PAN with SPP layer and DropBlock\cite{ghiasi2018dropblock} and the lightweight IoU aware head. In PP-YOLOv2, ReLU activation function is used in backbone while mish activation function is used in neck. Following YOLOv3, PP-YOLOv2 only assigns one anchor box for each ground truth object. In addition to classification loss, regression loss and objectness loss, PP-YOLOv2 also uses IoU loss and IoU aware loss to boost the performance. For more details, please refer to \cite{huang2021pp-yolov2}.

\subsection{Improvement of PP-YOLOE}




\noindent {\bf Anchor-free.} As mentioned above, PP-YOLOv2\cite{huang2021pp-yolov2} assigns ground truths in an anchor-based manner. However, anchor mechanism introduces a number of hyper-parameters and depends on hand-crafted design which may not generalize well on other datasets. For the above reason, we introduce anchor-free method in PP-YOLOv2. Following FCOS\cite{tian2019fcos}, which tiles one anchor point on each pixel, we set upper and lower bounds for three detection heads to assign ground truths to corresponding feature map. Then, the center of bounding box is calculated to select the closest pixel as positive samples. Following YOLO series, a 4D vector (x, y, w, h) is predicted for regression. This modification makes the model a little faster with the loss of 0.3 AP as shown in \Cref{roadmap}. Although upper and lower bounds are carefully set according to the anchor sizes of PP-YOLOv2, there are still some minor inconsistencies in the assignment results between anchor-based and anchor-free manner, which may lead to little precision drop.



~\\

\noindent {\bf Backbone and Neck.} Residual connections\cite{he2016deep, xie2017aggregated, hu2018squeeze} and dense connections\cite{huang2017densely, lee2019energy, rao2021treenet}  have been widely used in modern convolutional neural network. Residual connections introduce shortcut to relieve gradient vanishing problem and can be also regarded as a model ensemble approach. Dense connections aggregate intermediate features with diverse receptive fields, showing good performance on the object detection task. CSPNet\cite{wang2020cspnet} utilizes cross stage dense connections to lower computation burden without the loss of precision, which is popular among effective object detectors such as YOLOv5\cite{glenn_jocher_2022_6222936}, YOLOX\cite{ge2021yolox}. VoVNet\cite{lee2019energy} and subsequent TreeNet\cite{rao2021treenet} also show superior performance in object detection and instance segmentation. Inspired by these works, we propose a novel RepResBlock by combining the residual connections and dense connections, which is used in our backbone and neck.

Originating from TreeBlock\cite{rao2021treenet}, our RepResBlock is shown in \cref{fig:figure2b} during the training phase and \cref{fig:figure2c} during the inference phase. Firstly, we simplify the original TreeBlock (\cref{fig:figure2a}). Then, we replace the concatenation operation with element-wise add operation (\cref{fig:figure2b}), because of the approximation of these two operations to some extent shown in RMNet \cite{meng2021rmnet}. Thus, during the inference phase, we can re-parameterizes RepResBlock to a basic residual block (\cref{fig:figure2c}) used by ResNet-34 in a RepVGG\cite{ding2021repvgg} style. 

We use proposed RepResBlock to build backbone and neck. Similar to ResNet, our backbone, named CSPRepResNet, contains one stem composed of three convolution layer and four subsequent stages stacked by our RepResBlock as shown in \cref{fig:figure2d}. In each stage, cross stage partial connections are used to avoid numerous parameters and computation burden brought by lots of 3 $\times$ 3 convolution layers. ESE (Effective Squeeze and Extraction) layer is also used to impose channel attention in each CSPRepResStage while building backbone. We build neck with proposed RepResBlock and CSPRepResStage following PP-YOLOv2\cite{huang2021pp-yolov2}. Different from backbone, shortcut in RepResBlock and ESE layer in CSPRepResStage are removed in neck.

We use width multiplier $\alpha$ and depth multiplier $\beta$ to scale the basic backbone and neck jointly like YOLOv5\cite{glenn_jocher_2022_6222936}. Thus, we can get a series of detection network with different parameters and computation cost. The width setting of basic backbone is [64, 128, 256, 512, 1024]. Except for the stem, the depth setting of basic backbone is [3, 6, 6, 3]. The width setting and depth setting of basic neck are [192, 384, 768] and 3 respectively. \Cref{table1} shows the specification of width multiplier $\alpha$ and depth multiplier $\beta$ for different model. Such modifications obtains 0.7\% AP performance improvements -- 49.5\% AP as shown in \Cref{roadmap}. 
\begin{table}[!h]
	\centering
	\begin{center}
		\begin{tabular}{c|c|c}
			\hline
			    & width multiplier $\alpha$ & depth multiplier $\beta$ \\
			\hline
			
			s   & 0.50 & 0.33   \\
			m   & 0.75 & 0.67   \\
			l   & 1.00 & 1.00   \\
			x   & 1.25 & 1.33   \\
			\hline
		\end{tabular}
	\end{center}
	\caption{Width multiplier $\alpha$ and depth multiplier $\beta$ specification for a series of networks}
	\label{table1}
\end{table}


~\\

\begin{table*}[ht]
	\begin{center}
		\begin{tabular}{l|l|c|c|c|c}
			\hline
			Model & mAP(\%) & Parameters(M) & GFLOPs &  Latency(ms) & FPS \\
			\hline \hline
			
			PP-YOLOv2 baseline model & 49.1   & 54.58  & 115.77  & 14.5  & 68.9  \\
			\hline
			
			 +Anchor-free & 48.8 (\textcolor[RGB]{50,137,50}{\small\textbf{$-$0.3}}) & 54.27     & 114.78     &  14.3    & 69.8   \\
			 +CSPRepResNet & 49.5 (\textcolor[RGB]{225,10,10}{\small\textbf{$+$0.7}}) & 47.42     & 101.87     & 11.7    & 85.5   \\			
			 +TAL       & 50.4 (\textcolor[RGB]{225,10,10}{\small\textbf{$+$0.9}}) & 48.32     & 104.75     & 11.9	& 84.0    \\
			 +ET-head   &  \textbf{50.9} (\textcolor[RGB]{225,10,10}{\small\textbf{$+$0.5}}) & 52.20     & 110.07     & 12.8	 &  78.1  \\
			\hline
		\end{tabular}
	\end{center}
	
	\caption{Ablation study of PP-YOLOE-l on COCO \textit{val}. We use 640$\times$640 resolution as input with FP32-precision, and test on Tesla V100 without post-processing.}
	\label{roadmap}
\end{table*}

\noindent {\bf Task Alignment Learning (TAL).} To further improve the accuracy, label assignment is another aspect to be considered. YOLOX uses SimOTA as the label assignment strategy to improve performance. However, to further overcome the misalignment of classification and localization, task alignment learning (TAL) is proposed in TOOD\cite{tood}, which is composed of a dynamic label assignment and task aligned loss. Dynamic label assignment means prediction/loss aware. According to the prediction, it allocate dynamic number of positive anchors for each ground-truth. By explicitly aligning the two tasks, TAL can obtain the highest classification score and the most precise bounding box at the same time.
\begin{table}[!h]
	\centering
	\begin{center}
		\begin{tabular}{c|c}
			\hline
			Method    & mAP(0.5:0.95) \\
			\hline
			
			ATSS\cite{atss}       &  43.1 \\
			SimOTA\cite{ge2021yolox}  &  44.3 \\
			TAL\cite{tood}         &  \textbf{45.2} \\
			\hline
		\end{tabular}
	\end{center}
	\caption{Different label assignment on base model. We use CSPRepResStage as backbone and neck, one 1$\times$1 conv layer as head, and only train 36 epochs on COCO \textit{train2017}.}
	\label{label_assignment_ablation}
\end{table}

For task aligned loss, TOOD use a normalized $t$, namely $\hat{t}$, to replace the target in loss. It adopts the largest IoU within each instance as the normalization. The \textit{Binary Cross Entropy} (BCE) for the classification can be rewritten as:
\begin{equation}
	 L_{cls-pos}=\sum_{i=1}^{N_{pos}}BCE\left ( p_{i}, \hat{t}_i \right ) \label{tood_bce}
\end{equation}
We investigate the performance using different label assignment strategy. We conduct this experiment on above modified model, which use CSPRepResNet as backbone. For get the verification results quickly, we only train 36 epochs on COCO \textit{train2017} and verify it on COCO \textit{val}. As shown in \Cref{label_assignment_ablation}, TAL achieves the best 45.2\% AP performance. We use TAL to replace label assignment like FCOS style and achieve 0.9\% AP improvement -- 50.4\% AP as shown in \Cref{roadmap}.

~\\

\noindent {\bf Efficient Task-aligned Head (ET-head).} In object detection, the task conflict between classification and localization is a well-known problem. Corresponding solutions are proposed in many papers\cite{tood,zhang2021varifocalnet,li2020generalized,2021Rethinking}. YOLOX's decoupled head draws lessons from most of the one-stage and two-stage detectors, and successfully apply to YOLO model to improve accuracy. However, the decoupled head may make the classification and localization tasks separate and independent, and lack of task specific learning. Based on TOOD\cite{tood}, we improve the head and propose ET-head with the goal of both speed and accuracy. As shown in \cref{head_fig}, we use ESE to replace the layer attention in TOOD, simplify the alignment of classification branches to shortcut, and replace the alignment of regression branches with distribution focal loss (DFL) layer\cite{li2020generalized}. Through the above changes, the ET-head brings an increase of 0.9ms on V100.

For the learning of classification and location tasks, we choose varifocal loss (VFL) and distribution focal loss (DFL) respectively. PP-Picodet\cite{picodet} successfully applys VFL and DFL in object detectors, and obtains performance improvement. For VFL in \cite{zhang2021varifocalnet}, different from the quality focal loss (QFL) in \cite{li2020generalized}, VFL uses target score to weight the loss of positive samples. This implementation makes the contribution of positive samples with high IoU to loss relatively large. This also makes the model pay more attention to high-quality samples rather than those low-quality ones at training time. The same is that both use IoU-aware classification score (IACS) as the target to predict. This can effectively learn a joint representation of classification score and localization quality estimation, which enables high consistency between training and inference. For DFL, in order to solve the problem of inflexible representation of bounding box, \cite{li2020generalized} proposes to use general distribution to predict bounding box. Our model is supervised by the loss function:
\begin{equation}
	 Loss=\frac{\alpha \cdot loss_{VFL}+ \beta \cdot loss_{GIoU}+ \gamma \cdot loss_{DFL}}{\sum_{i}^{N_{pos}}\hat{t}}
\end{equation}
where $\hat{t}$ denote the normalized target score, see \cref{tood_bce}. And as shown in \Cref{roadmap}, the ET-head obtains 0.5\% AP improvement -- 50.9\% AP.



\section{Experiment}


In this section, we present the experiments details and results. All experiments are trained on
MS COCO-2017 training set with 80 classes and 118k images. For ablation study, we use the standard COCO AP metric with single scale on MS COCO-2017 validation set with 5000 images. And we report final results using MS COCO-2017 \textit{test-dev}.

\subsection{Implementation details}

We use stochastic gradient descent (SGD) with momentum = 0.9 and weight decay = 5e-4. We use cosine learning rate schedule, total epochs are 300, warmup epochs are 5, and base learning rate is 0.01. The total batch size is 64 on $8\times32$ G V100 GPU devices by default, and we follow linear scaling rule\cite{linear-scale-rule} to adjust learning rate. The exponential moving average (EMA) strategy with decay = 0.9998 is also adopted during training process. We only use some basic data augmentations, including random crop, random horizontal flip, color distortion, and multi-scale. Specially, input size is evenly drawn from 320 to 768 with 32 stride.

\begin{table*}[ht]
		\centering
		\resizebox{1.0\textwidth}{!}{
			\begin{tabular}{l|l|c|cc|cccccc}
				\hline
				\multirow{2}{*}{\textbf{Method}} & \multirow{2}{*}{\textbf{Backbone}} & \multirow{2}{*}{\textbf{Size}} &\multicolumn{2}{c|}{\textbf{FPS} (v100)} &
				\multirow{2}{*}{\textbf{AP}} & \multirow{2}{*}{\textbf{AP$_{50}$}} & \multirow{2}{*}{\textbf{AP$_{75}$}} & \multirow{2}{*}{\textbf{AP$_S$}} & \multirow{2}{*}{\textbf{AP$_M$}} & \multirow{2}{*}{\textbf{AP$_L$}}\\	
				\cline{4-5} 
				& & & \textbf{w/o TRT} & \textbf{with TRT} & & & & & &\\			
				\hline
				YOLOv3 + ASFF* \cite{Liu2019Learning} & Darknet-53 & 320 &  60   &- & 38.1\% & 57.4\% & 42.1\% & 16.1\% & 41.6\% & 53.6\% \\
				YOLOv3 + ASFF* \cite{Liu2019Learning}& Darknet-53 & 416 &  54   & - &40.6\% & 60.6\% & 45.1\% & 20.3\% & 44.2\% & 54.1\% \\

				YOLOv4~\cite{bochkovskiy2020yolov4} & CSPDarknet-53 & 416 & 96  & - & 41.2\% & 62.8\% & 44.3\% & 20.4\% & 44.4\% & 56.0\% \\
				YOLOv4~\cite{bochkovskiy2020yolov4} & CSPDarknet-53 & 512 & 83  & -  & 43.0\% & 64.9\% & 46.5\% & 24.3\% & 46.1\% & 55.2\% \\
				
				YOLOv4-CSP~\cite{wang2021scaledyolov4} & Modified CSPDarknet-53 & 512 & 97   & - & 46.2\%  & 64.8\% & 50.2\% & 24.6\% & 50.4\% & 61.9\% \\
				YOLOv4-CSP~\cite{wang2021scaledyolov4} &  Modified CSPDarknet-53 & 640 & 73  &  - & 47.5\% & 66.2\% & 51.7\% & 28.2\% & 51.2\% & 59.8\% \\
				
				\hline
				EfficientDet-D0~\cite{tan2020efficientdet} & Efficient-B0  & 512 &  98.0 & - & 33.8\% & 52.2\% & 35.8\% & 12.0\% & 38.3\% & 51.2\% \\
				EfficientDet-D1~\cite{tan2020efficientdet} & Efficient-B1 & 640 &  74.1  &- & 39.6\% & 58.6\% & 42.3\% & 17.9\% & 44.3\% & 56.0\% \\
				EfficientDet-D2~\cite{tan2020efficientdet} & Efficient-B2 & 768 & 56.5  &- & 43.0\% & 62.3\% & 46.2\% & 22.5\% & 47.0\% & 58.4\% \\
				EfficientDet-D2~\cite{tan2020efficientdet} & Efficient-B3 & 896 & 34.5  &- & 45.8\% & 65.0\% & 49.3\% & 26.6\% & 49.4\% & 59.8\% \\
				
				
				
				\hline
				PP-YOLO~\cite{long2020pp} & ResNet50-vd-dcn & 320 & 132.2$^+$ &  242.2$^+$ & 39.3\% & 59.3\% & 42.7\% & 16.7\% & 41.4\% & 57.8\% \\
				PP-YOLO~\cite{long2020pp} & ResNet50-vd-dcn & 416 & 109.6$^+$ & 215.4$^+$ & 42.5\% & 62.8\% & 46.5\% & 21.2\% & 45.2\% & 58.2\% \\
				PP-YOLO~\cite{long2020pp} & ResNet50-vd-dcn & 512 & 89.9$^+$ & 188.4$^+$ & 44.4\%  & 64.6\% & 48.8\% & 24.4\% & 47.1\% & 58.2\% \\
				PP-YOLO~\cite{long2020pp}& ResNet50-vd-dcn & 608 & 72.9$^+$ &  155.6$^+$ & 45.9\% & 65.2\% & 49.9\% & 26.3\% & 47.8\% & 57.2\% \\
				
				\hline
				PP-YOLOv2 \cite{huang2021pp-yolov2} & ResNet50-vd-dcn & 320 & 123.3 & 152.9 & 43.1\% & 61.7\% & 46.5\% & 19.7\% & 46.3\% & 61.8\% \\
				PP-YOLOv2 \cite{huang2021pp-yolov2} & ResNet50-vd-dcn & 416 & 102$^+$ & 145.1$^+$ & 46.3\% & 65.1\% & 50.3\% & 23.9\% & 50.2\% & 62.2\% \\
				PP-YOLOv2 \cite{huang2021pp-yolov2} & ResNet50-vd-dcn & 512 & 93.4$^+$ &  141.2$^+$ & 48.2\% & 67.1\% & 52.7\% & 27.7\% & 52.1\% & 62.1\% \\
				PP-YOLOv2 \cite{huang2021pp-yolov2} & ResNet50-vd-dcn & 640 & 68.9$^+$ & 106.5$^+$ & 49.5\% & 68.2\% & 54.4\% & 30.7\% & 52.9\% & 61.2\% \\
				PP-YOLOv2 \cite{huang2021pp-yolov2} & ResNet101-vd-dcn & 640 & 50.3$^+$ &  87.0$^+$ & 50.3\% & 69.0\% & 55.3\% & 31.6\% & 53.9\% & 62.4\% \\
				
				\hline
				YOLOv5-s \cite{glenn_jocher_2022_6222936} & Modified CSP v6 & 640 & 156.2$^+$ & 454.5$^*$ & 37.4\% & 56.8\% & - & - & - & - \\
				YOLOv5-m \cite{glenn_jocher_2022_6222936} & Modified CSP v6 & 640 & 121.9$^+$ & 263.1$^*$ & 45.4\% & 64.1\% & - & - & - & - \\
				YOLOv5-l \cite{glenn_jocher_2022_6222936} & Modified CSP v6 & 640 & 99.0$^+$ & 172.4$^*$ & 49.0\%  & 67.3\% & - & - & - & - \\
				YOLOv5-x \cite{glenn_jocher_2022_6222936} & Modified CSP v6 & 640 & 82.6$^+$ & 117.6$^*$ & 50.7\% & 68.9\% & - & - & - & - \\
				
				
				\hline
				YOLOX-s \cite{ge2021yolox} & Modified CSP v5 & 640 & 119.0$^*$ $\vert$ 102.0$^+$  & 246.9$^*$ & 40.5\% & - & - & - & - & - \\
				YOLOX-m \cite{ge2021yolox}& Modified CSP v5 & 640 & 96.1$^*$ $\vert$ 81.3$^+$ & 177.3$^*$ & 47.2\% & - & - & - & - & - \\
				YOLOX-l \cite{ge2021yolox}& Modified CSP v5 & 640 & 62.5$^*$ $\vert$ 68.9$^+$  & 120.1$^*$ & 50.1\% & - & - & - & - & - \\
				YOLOX-x \cite{ge2021yolox}& Modified CSP v5 & 640 & 40.3$^*$ $\vert$ 57.8$^+$ & 87.4$^*$  & 51.5\% & - & - & - & - & - \\
				
				\hline
			    \hline
				\textbf{PP-YOLOE-s} & CSPRepResNet & 640 & 208.3 & 333.3 & \textbf{43.1\%} & \textbf{60.5\%} & \textbf{46.6\%} & \textbf{23.2\%} & \textbf{46.4\%} & \textbf{56.9\%} \\
				\textbf{PP-YOLOE-m} & CSPRepResNet & 640 & 123.4 & 208.3 & \textbf{48.9\%} & \textbf{66.5\%} & \textbf{53.0\%} & \textbf{28.6\%} & \textbf{52.9\%} & \textbf{63.8\%} \\
				\textbf{PP-YOLOE-l} & CSPRepResNet & 640 & 78.1 &  149.2 & \textbf{51.4\%} & \textbf{68.9\%} & \textbf{55.6\%} & \textbf{31.4\%} & \textbf{55.3\%} & \textbf{66.1\%} \\
				\textbf{PP-YOLOE-x} & CSPRepResNet & 640 & 45.0 &  95.2 & \textbf{52.2\%} & \textbf{69.9\%} & \textbf{56.5\%} & \textbf{33.3\%} & \textbf{56.3\%} & \textbf{66.4\%} \\
				\hline
				
				\hline
				\hline
				\textbf{PP-YOLOE+-s} & CSPRepResNet & 640 & 208.3 & 333.3 & \textbf{43.7\%} & \textbf{60.6\%} & \textbf{47.9\%} & \textbf{26.5\%} & \textbf{47.5\%} & \textbf{59.0\%} \\
				\textbf{PP-YOLOE+-m} & CSPRepResNet & 640 & 123.4 & 208.3 & \textbf{49.8\%} & \textbf{67.1\%} & \textbf{54.5\%} & \textbf{31.8\%} & \textbf{53.9\%} & \textbf{66.2\%} \\
				\textbf{PP-YOLOE+-l} & CSPRepResNet & 640 & 78.1 &  149.2 & \textbf{52.9\%} & \textbf{70.1\%} & \textbf{57.9\%} & \textbf{35.2\%} & \textbf{57.5\%} & \textbf{69.1\%} \\
				\textbf{PP-YOLOE+-x} & CSPRepResNet & 640 & 45.0 &  95.2 & \textbf{54.7\%} & \textbf{72.0\%} & \textbf{59.9\%} & \textbf{37.9\%} & \textbf{59.3\%} & \textbf{70.4\%} \\
				\hline
			\end{tabular}
		}
		\vspace{0.1cm}
		\caption{Comparison of the speed and accuracy of different object detectors on COCO 2017 \textit{test-dev}. Results marked by "+" are updated results from the corresponding official release. Results marked by "*" are tested in our environment using official codebase and model. The input size of YOLOv5 is not exactly square of $640\times640$ in validation and speed test, so we skip it in the table. The default precision of speed is FP32 for \textit{w/o trt} and FP16 for \textit{with trt}. Moreover, we provide both FP32 and FP16 for YOLOX \textit{w/o trt} scene, the FP32 speed on the left side of split line and FP16 speed on the right. PP-YOLOE+ uses the model pre trained on the Objects365\cite{2020Objects365} datasets.
		}
		\label{tab-state-of-the-art}
\end{table*}






\subsection{Comparsion with Other SOTA Detectors}

\Cref{tab-state-of-the-art} and \Cref{figure1} show comparison of the results on MS-COCO test split with other state-of-the-art object detectors. We re-evaluate YOLOv5\cite{glenn_jocher_2022_6222936} and YOLOX\cite{ge2021yolox} using official codebase because they have non-scheduled updates. We compare model inference speed with batch size = 1 (without data preprocess and non-maximum suppression). However, PP-YOLOE series using paddle inference engine. Further, for fair comparison, we also test the FP16 precision speed based on tensorRT 6.0 in the same environment. It should be emphasized that PaddlePaddle\footnote{\scriptsize\url{https://github.com/PaddlePaddle/Paddle}\label{paddle}} officially supports tensorRT for model deployment. Therefore, PP-YOLOE can use paddle inference with tensorRT directly, and other tests follow the official guidelines.



\section{Conclusion}
In this report, we present several updates to PP-YOLOv2, including scalable backbone-neck architecture, efficient task aligned head, advanced label assignment strategy and refined objective loss function, which forms a series high-performance object detectors called PP-YOLOE. Meanwhile, we present s/m/l/x models which can cover different scenarios in practice. Moreover, these models can smoothly transition to deployment, with PaddlePaddle official support. We hope these designs with encouraging results can provide inspirations for developers and researchers.




{\small
\bibliographystyle{ieee_fullname}
\bibliography{egbib}

\begin{thebibliography}{10}\itemsep=-1pt

\bibitem{ppdet2021}
PaddlePaddle Authors.
\newblock {PaddleDetection}, object detection and instance segmentation toolkit
  based on paddlepaddle.
\newblock \url{https://github.com/PaddlePaddle/PaddleDetection}, 2021.

\bibitem{bochkovskiy2020yolov4}
Alexey Bochkovskiy, Chien-Yao Wang, and Hong-Yuan~Mark Liao.
\newblock Yolov4: Optimal speed and accuracy of object detection.
\newblock {\em arXiv preprint arXiv:2004.10934}, 2020.

\bibitem{dai2017deformable}
Jifeng Dai, Haozhi Qi, Yuwen Xiong, Yi Li, Guodong Zhang, Han Hu, and Yichen
  Wei.
\newblock Deformable convolutional networks.
\newblock In {\em Proceedings of the IEEE international conference on computer
  vision}, pages 764--773, 2017.

\bibitem{ding2021repvgg}
Xiaohan Ding, Xiangyu Zhang, Ningning Ma, Jungong Han, Guiguang Ding, and Jian
  Sun.
\newblock Repvgg: Making vgg-style convnets great again.
\newblock In {\em Proceedings of the IEEE/CVF Conference on Computer Vision and
  Pattern Recognition}, pages 13733--13742, 2021.

\bibitem{tood}
Chengjian Feng, Yujie Zhong, Yu Gao, Matthew~R Scott, and Weilin Huang.
\newblock Tood: Task-aligned one-stage object detection.
\newblock In {\em Proceedings of the IEEE/CVF International Conference on
  Computer Vision}, pages 3510--3519, 2021.

\bibitem{ge2021yolox}
Zheng Ge, Songtao Liu, Feng Wang, Zeming Li, and Jian Sun.
\newblock Yolox: Exceeding yolo series in 2021.
\newblock {\em arXiv preprint arXiv:2107.08430}, 2021.

\bibitem{ghiasi2018dropblock}
Golnaz Ghiasi, Tsung-Yi Lin, and Quoc~V Le.
\newblock Dropblock: A regularization method for convolutional networks.
\newblock {\em Advances in neural information processing systems}, 31, 2018.

\bibitem{linear-scale-rule}
Priya Goyal, Piotr Doll{\'{a}}r, Ross~B. Girshick, Pieter Noordhuis, Lukasz
  Wesolowski, Aapo Kyrola, Andrew Tulloch, Yangqing Jia, and Kaiming He.
\newblock Accurate, large minibatch {SGD:} training imagenet in 1 hour.
\newblock {\em CoRR}, abs/1706.02677, 2017.

\bibitem{he2016deep}
Kaiming He, Xiangyu Zhang, Shaoqing Ren, and Jian Sun.
\newblock Deep residual learning for image recognition.
\newblock In {\em Proceedings of the IEEE conference on computer vision and
  pattern recognition}, pages 770--778, 2016.

\bibitem{he2019bag}
Tong He, Zhi Zhang, Hang Zhang, Zhongyue Zhang, Junyuan Xie, and Mu Li.
\newblock Bag of tricks for image classification with convolutional neural
  networks.
\newblock In {\em Proceedings of the IEEE/CVF Conference on Computer Vision and
  Pattern Recognition}, pages 558--567, 2019.

\bibitem{hu2018squeeze}
Jie Hu, Li Shen, and Gang Sun.
\newblock Squeeze-and-excitation networks.
\newblock In {\em Proceedings of the IEEE conference on computer vision and
  pattern recognition}, pages 7132--7141, 2018.

\bibitem{huang2017densely}
Gao Huang, Zhuang Liu, Laurens Van Der~Maaten, and Kilian~Q Weinberger.
\newblock Densely connected convolutional networks.
\newblock In {\em Proceedings of the IEEE conference on computer vision and
  pattern recognition}, pages 4700--4708, 2017.

\bibitem{huang2021pp-yolov2}
Xin Huang, Xinxin Wang, Wenyu Lv, Xiaying Bai, Xiang Long, Kaipeng Deng,
  Qingqing Dang, Shumin Han, Qiwen Liu, Xiaoguang Hu, Dianhai Yu, Yanjun Ma,
  and Osamu Yoshie.
\newblock Pp-yolov2: A practical object detector, 2021.

\bibitem{glenn_jocher_2022_6222936}
Glenn Jocher, Ayush Chaurasia, Alex Stoken, Jirka Borovec, NanoCode012, Yonghye
  Kwon, TaoXie, Jiacong Fang, imyhxy, Kalen Michael, Lorna, Abhiram V, Diego
  Montes, Jebastin Nadar, Laughing, tkianai, yxNONG, Piotr Skalski, Zhiqiang
  Wang, Adam Hogan, Cristi Fati, Lorenzo Mammana, AlexWang1900, Deep Patel,
  Ding Yiwei, Felix You, Jan Hajek, Laurentiu Diaconu, and Mai~Thanh Minh.
\newblock {ultralytics/yolov5: v6.1 - TensorRT, TensorFlow Edge TPU and
  OpenVINO Export and Inference}, Feb. 2022.

\bibitem{lee2019energy}
Youngwan Lee, Joong-won Hwang, Sangrok Lee, Yuseok Bae, and Jongyoul Park.
\newblock An energy and gpu-computation efficient backbone network for
  real-time object detection.
\newblock In {\em Proceedings of the IEEE/CVF Conference on Computer Vision and
  Pattern Recognition Workshops}, pages 0--0, 2019.

\bibitem{li2020generalized}
Xiang Li, Wenhai Wang, Lijun Wu, Shuo Chen, Xiaolin Hu, Jun Li, Jinhui Tang,
  and Jian Yang.
\newblock Generalized focal loss: Learning qualified and distributed bounding
  boxes for dense object detection.
\newblock {\em arXiv preprint arXiv:2006.04388}, 2020.

\bibitem{Liu2019Learning}
Songtao Liu, Di Huang, and Yunhong Wang.
\newblock Learning spatial fusion for single-shot object detection, 2019.

\bibitem{long2020pp}
Xiang Long, Kaipeng Deng, Guanzhong Wang, Yang Zhang, Qingqing Dang, Yuan Gao,
  Hui Shen, Jianguo Ren, Shumin Han, Errui Ding, and Shilei Wen.
\newblock Pp-yolo: An effective and efficient implementation of object
  detector.
\newblock {\em arXiv preprint arXiv:2007.12099}, 2020.

\bibitem{meng2021rmnet}
Fanxu Meng, Hao Cheng, Jiaxin Zhuang, Ke Li, and Xing Sun.
\newblock Rmnet: Equivalently removing residual connection from networks.
\newblock {\em arXiv preprint arXiv:2111.00687}, 2021.

\bibitem{rao2021treenet}
Lu Rao.
\newblock Treenet: A lightweight one-shot aggregation convolutional network.
\newblock {\em arXiv preprint arXiv:2109.12342}, 2021.

\bibitem{redmon2016you}
Joseph Redmon, Santosh Divvala, Ross Girshick, and Ali Farhadi.
\newblock You only look once: Unified, real-time object detection.
\newblock In {\em Proceedings of the IEEE conference on computer vision and
  pattern recognition}, pages 779--788, 2016.

\bibitem{redmon2017yolo9000}
Joseph Redmon and Ali Farhadi.
\newblock Yolo9000: better, faster, stronger.
\newblock In {\em Proceedings of the IEEE conference on computer vision and
  pattern recognition}, pages 7263--7271, 2017.

\bibitem{redmon2018yolov3}
Joseph Redmon and Ali Farhadi.
\newblock Yolov3: An incremental improvement.
\newblock {\em arXiv preprint arXiv:1804.02767}, 2018.

\bibitem{2020Objects365}
S. Shao, Z. Li, T. Zhang, C. Peng, and J. Sun.
\newblock Objects365: A large-scale, high-quality dataset for object detection.
\newblock In {\em 2019 IEEE/CVF International Conference on Computer Vision
  (ICCV)}, 2020.

\bibitem{tan2020efficientdet}
Mingxing Tan, Ruoming Pang, and Quoc~V. Le.
\newblock Efficientdet: Scalable and efficient object detection, 2020.

\bibitem{tian2019fcos}
Zhi Tian, Chunhua Shen, Hao Chen, and Tong He.
\newblock Fcos: Fully convolutional one-stage object detection.
\newblock In {\em Proceedings of the IEEE/CVF international conference on
  computer vision}, pages 9627--9636, 2019.

\bibitem{wang2021scaledyolov4}
Chien-Yao Wang, Alexey Bochkovskiy, and Hong-Yuan~Mark Liao.
\newblock Scaled-yolov4: Scaling cross stage partial network, 2021.

\bibitem{wang2020cspnet}
Chien-Yao Wang, Hong-Yuan~Mark Liao, Yueh-Hua Wu, Ping-Yang Chen, Jun-Wei
  Hsieh, and I-Hau Yeh.
\newblock Cspnet: A new backbone that can enhance learning capability of cnn.
\newblock In {\em Proceedings of the IEEE/CVF conference on computer vision and
  pattern recognition workshops}, pages 390--391, 2020.

\bibitem{wang2020solov2}
Xinlong Wang, Rufeng Zhang, Tao Kong, Lei Li, and Chunhua Shen.
\newblock Solov2: Dynamic and fast instance segmentation.
\newblock {\em Advances in Neural information processing systems},
  33:17721--17732, 2020.

\bibitem{xie2017aggregated}
Saining Xie, Ross Girshick, Piotr Doll{\'a}r, Zhuowen Tu, and Kaiming He.
\newblock Aggregated residual transformations for deep neural networks.
\newblock In {\em Proceedings of the IEEE conference on computer vision and
  pattern recognition}, pages 1492--1500, 2017.

\bibitem{2021Rethinking}
Yang Yang, Min Li, Bo Meng, Junxing Ren, Degang Sun, and Zihao Huang.
\newblock Rethinking the aligned and misaligned features in one-stage object
  detection.
\newblock 2021.

\bibitem{picodet}
Guanghua Yu, Qinyao Chang, Wenyu Lv, Chang Xu, Cheng Cui, Wei Ji, Qingqing
  Dang, Kaipeng Deng, Guanzhong Wang, Yuning Du, Baohua Lai, Qiwen Liu,
  Xiaoguang Hu, Dianhai Yu, and Yanjun Ma.
\newblock Pp-picodet: {A} better real-time object detector on mobile devices.
\newblock {\em CoRR}, abs/2111.00902, 2021.

\bibitem{zhang2021varifocalnet}
Haoyang Zhang, Ying Wang, Feras Dayoub, and Niko Sunderhauf.
\newblock Varifocalnet: An iou-aware dense object detector.
\newblock In {\em Proceedings of the IEEE/CVF Conference on Computer Vision and
  Pattern Recognition}, pages 8514--8523, 2021.

\bibitem{atss}
Shifeng Zhang, Cheng Chi, Yongqiang Yao, Zhen Lei, and Stan~Z Li.
\newblock Bridging the gap between anchor-based and anchor-free detection via
  adaptive training sample selection.
\newblock In {\em Proceedings of the IEEE/CVF conference on computer vision and
  pattern recognition}, pages 9759--9768, 2020.

\bibitem{zhu2018deformable}
Xizhou Zhu, Han Hu, Stephen Lin, and Jifeng Dai.
\newblock Deformable convnets v2: More deformable.
\newblock {\em Better Results}, 2018.

\end{thebibliography}
}

\end{document}